\documentclass[letterpaper, 10 pt, conference]{ieeeconf}  
\makeatletter
\let\NAT@parse\undefined
\makeatother
\usepackage{hyperref}
\usepackage[T1]{fontenc}
\usepackage{graphicx}
\usepackage{algorithm}
\usepackage{algorithmic}
\usepackage{mathrsfs}
\usepackage{amssymb}
\usepackage{amsmath} 
\usepackage{multirow} 
\usepackage{booktabs}
\usepackage{color,xcolor,colortbl}

\usepackage{newfloat}
\usepackage{listings}

\IEEEoverridecommandlockouts                              

\title{\LARGE \bf
Cognition-Inspired Dual-Stream Semantic Enhancement for Vision-Based Dynamic Emotion Modeling
}

\author{Huanzhen Wang\textsuperscript{1}, Ziheng Zhou\textsuperscript{1}, Zeng Tao\textsuperscript{1}, Aoxing Li\textsuperscript{1}, Yingkai Zhao\textsuperscript{1}, Yuxuan Lin\textsuperscript{2}, Yan Wang\textsuperscript{3,*},  \\ Wenqiang Zhang\textsuperscript{1,2,*} 
\thanks{$^{1}$Huanzhen Wang, Ziheng Zhou, Zeng Tao, Aoxing Li and Yingkai Zhao are with College of Intelligent Robotics and Advanced Manufacturing, Fudan University, Shanghai, China. {\tt\small {\{hzwang24, zhouzh24, axli24, ykzhao24\}@m.fudan.edu.cn}, {ztao19@fudan.edu.cn}}}%
\thanks{$^{2}$Yuxuan Lin and Wenqiang Zhang are with Shanghai Key Lab of Intelligent Information Processing, College of Computer Science and Artificial Intelligence, Fudan University, Shanghai, China. {\tt\small {yuxuanlin24@m.fudan.edu.cn}, {wqzhang@fudan.edu.cn}}}%
\thanks{$^{3}$Yan Wang is with School of Data Science and Engineering, East China Normal University, Shanghai, China. {\tt\small yanwang@dase.ecnu.edu.cn}}%
\thanks{*Corresponding author: Wenqiang Zhang and Yan Wang.}
}

\begin{document}

\maketitle
\thispagestyle{empty}
\pagestyle{empty}

\begin{abstract}
The human brain constructs emotional percepts not by processing facial expressions in isolation, but through a dynamic, hierarchical integration of sensory input with semantic and contextual knowledge. However, existing vision-based dynamic emotion modeling approaches often neglect emotion perception and cognitive theories. To bridge this gap between machine and human emotion perception, we propose cognition-inspired Dual-stream Semantic Enhancement (DuSE). Our model instantiates a dual-stream cognitive architecture. The first stream, a Hierarchical Temporal Prompt Cluster (HTPC), operationalizes the cognitive priming effect. It simulates how linguistic cues pre-sensitize neural pathways, modulating the processing of incoming visual stimuli by aligning textual semantics with fine-grained temporal features of facial dynamics. The second stream, a Latent Semantic Emotion Aggregator (LSEA), computationally models the knowledge integration process, akin to the mechanism described by the Conceptual Act Theory. It aggregates sensory inputs and synthesizes them with learned conceptual knowledge, reflecting the role of the hippocampus and default mode network in constructing a coherent emotional experience.  By explicitly modeling these neuro-cognitive mechanisms, DuSE provides a more neurally plausible and robust framework for dynamic facial expression recognition (DFER). Extensive experiments on challenging in-the-wild benchmarks validate our cognition-centric approach, demonstrating that emulating the brain's strategies for emotion processing yields state-of-the-art performance and enhances model interpretability.

\end{abstract}

\section{INTRODUCTION}

Facial expressions serve as the core cue for conveying human emotions, capable of rapidly activating emotion-processing brain regions such as the amygdala to enable emotional salience detection and empathy inference \cite{adolphs2002neural}. They also serve as a universal channel for emotional communication and are widely applied in fields such as healthcare, robotics, and human-computer interaction \cite{hossain2017emotion}. While dynamic facial expression recognition (DFER) leverages temporal cues for improved emotional interpretation \cite{zhao2021learning}, unimodal approaches often falter under occlusion or noise due to their limited perceptual scope \cite{ezzameli2023emotion}. To enhance robustness, recent research has turned to multimodal strategies integrating visual and auditory signals \cite{mai2024all}, with multiphysical methods emerging as a promising direction \cite{wang2022systematic}. The emergence of multimodal and cross-modal approaches also poses interpretability challenges for vision-based emotion modeling that align with human cognition. This paper is dedicated to exploring the design of bionic algorithm models that align with the theoretical foundations and practical requirements of affective cognitive science.

\begin{figure}[t]
  \centering
  \includegraphics[width=0.9\linewidth]{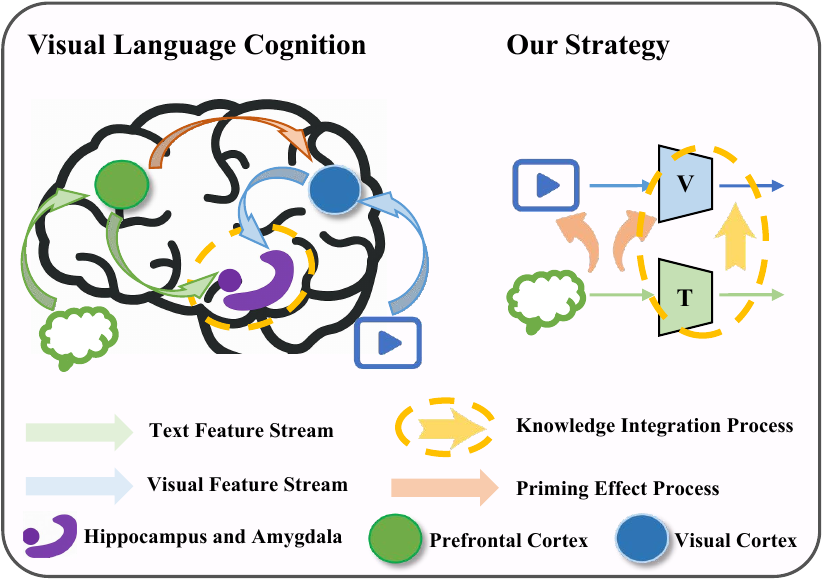}
  \caption{The priming effect and knowledge integration mechanisms in human emotional cognition inspire us to explore the progressive complementary relationship between prompts and knowledge in visual language sentiment modeling.}
  \label{mot-0}
\end{figure}

When the brain recognizes emotions, it integrates inputs from multiple sources such as vision and language, relying on core regions including the amygdala, insula, and prefrontal cortex \cite{gundem2022neurobiological}. Research indicates that emotional stimuli—whether visual facial expressions or auditory cues—activate regions including the amygdala and insula \cite{gerdes2014emotional}. Several key points warrant attention in the process of emotional transmission. On one hand, the brain exhibits a hierarchical structure across temporal scales. Experimental evidence indicates that lower-level sensory cortices possess intrinsic short time scales, whereas higher-level cross-modal networks operate on longer time scales \cite{golesorkhi2021brain}. On the other hand, the brain's perception of emotion is profoundly influenced by natural language and contextual semantics. Research reveals that when individuals receive information through narrative or linguistic channels, the hippocampus and default mode network rapidly retrieve relevant conceptual memories to interpret emotional cues  \cite{camacho2025cognitive}. Subsequently, the hippocampus and medial temporal lobe integrate these experiences into conceptual knowledge, which is then accessed through the default mode network. As shown in the Fig. \ref{mot-0}, these theories inspire us to explore the relationship between temporality and semantics, as well as between cues and knowledge, within the process of emotional modeling. 

\begin{figure}[t]
  \centering
  \includegraphics[width=0.9\linewidth]{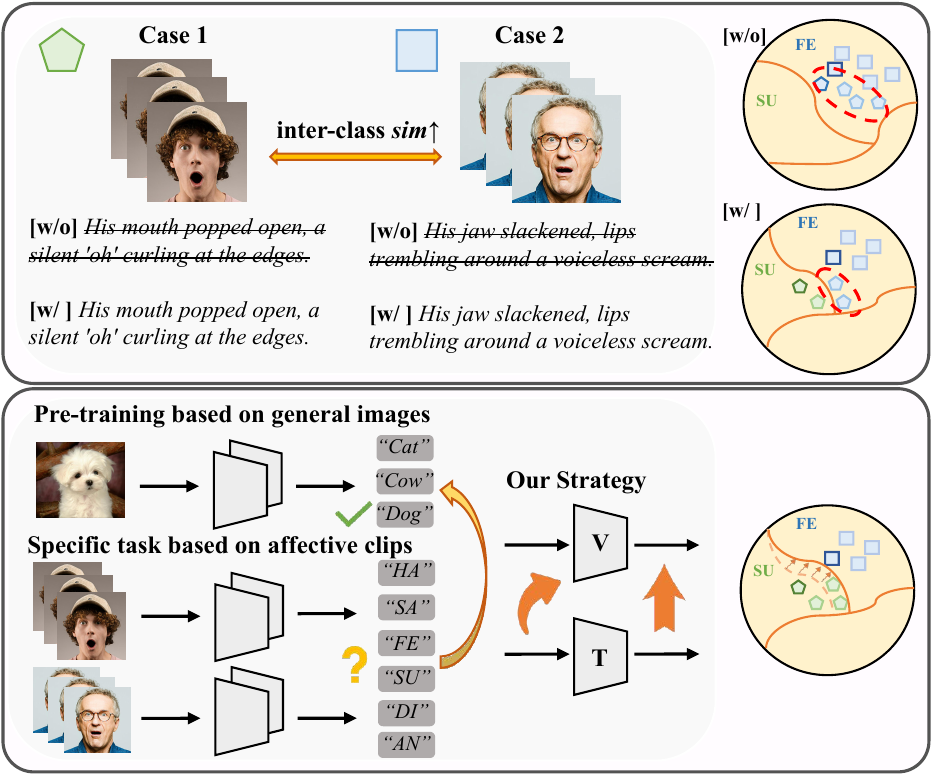}
  \caption{Supervision from natural language can effectively alleviate the problem of difficulty in classifying expressions with inter-class similarities. Our method aims to bridge the gap between general image-based pre-training models and specialized recognition tasks based on dynamic expression sequences.}
  \label{mot-1}
\end{figure}

In response to actual task requirements, as illustrated in Fig. \ref{mot-1}, subtle inter-class similarities and intra-class variations complicate DFER. Existing methods—whether supervised or self-supervised—typically rely on visual cues alone \cite{wang2024survey}, overlooking the semantic depth offered by language. One-hot labels lack contextual information, limiting generalization and interpretability. In contrast, natural language supervision introduces richer, human-aligned guidance that helps disambiguate similar expressions. Vision-language models like CLIP \cite{radford2021learning} have been adapted to DFER via prompt tuning and fine-tuning \cite{li2024cliper}, showing promising results.

Although CLIP exhibits strong generalization to novel concepts, its large-scale architecture and the scarcity of task-specific data render full-model fine-tuning impractical for downstream DFER tasks. While prior work in DFER has largely focused on adapting the visual encoder, findings from semantic segmentation \cite{zhang2024exploring} indicate that CLIP’s text encoder contains rich semantic priors that remain underexploited in emotion recognition. Limited modal interaction, challenging knowledge transfer, and interpretability constraints aligned with human emotional cognition restrict the potential of visual-language models in dynamic emotion modeling.           

To address the aforementioned challenges, we propose DuSE, a cognition-inspired dual-stream semantic enhancement framework for vision-based dynamic emotion modeling. Specifically, DuSE integrates a cross-modal prompt streaming to align textual emotion descriptions with fine-grained facial features, and a cross-domain knowledge streaming to transfer general visual knowledge to the facial expression domain. The algorithmic implementation of this “dual-mechanism” framework—which integrates pre-set expectations and knowledge through prompts—enables embodied cross-modal emotion perception.

Our main contributions are summarized as follows:

\begin{itemize}
\item Through cognitive affective analysis, we have revealed the gap between human multimodal dynamic emotion perception and unimodal DFER systems. Integrating cognitive theory with current applications, we propose the DuSE method.

\item Inspired by the priming effect and knowledge integration process in cognitive theory, we design the Hierarchical Temporal Prompt Cluster (HTPC) to support cross-modal prompt streaming and we design the Latent Semantic Emotion Aggregator (LSEA) to support cross-domain knowledge streaming.

\item We validate the effectiveness of our approach through extensive experiments on two challenging in-the-wild DFER benchmark datasets, demonstrating its superior performance over state-of-the-art methods.
\end{itemize}

\begin{figure*}[t]
  \centering
  \includegraphics[width=1\linewidth]{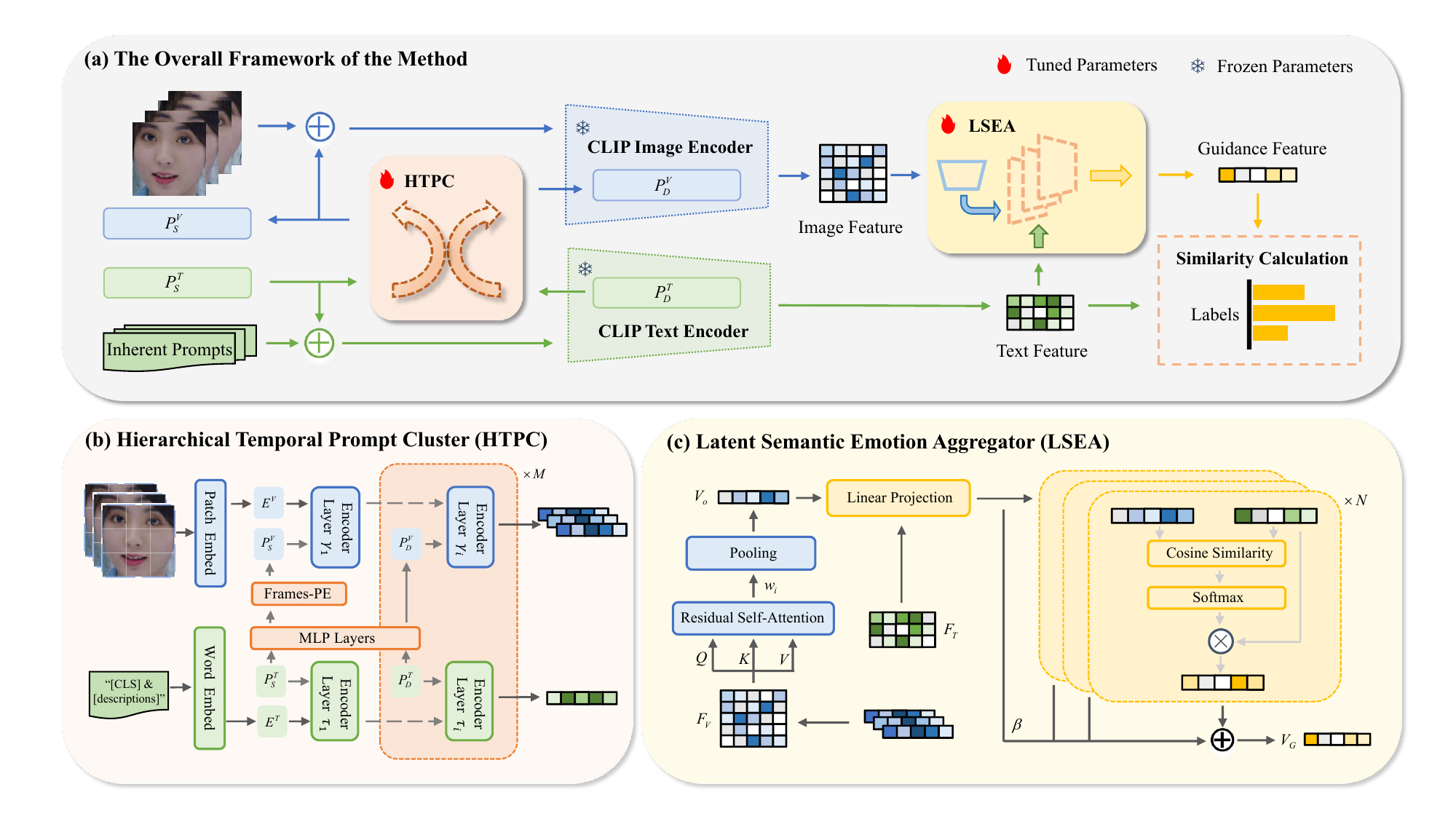}
  \caption{\textbf{Overall architecture of DuSE}. (a) shows the overall methodological framework. (b) shows HTPC, which contributes to the cross-modal prompt streaming. (c) shows LSEA, which contributes to cross-domain knowledge streaming.}
  \label{pip}
\end{figure*}

\section{RELATED WORKS}
\subsection{Cognitive Science and Neuroscience}
Advances in cognitive science and neuroscience provide theoretical foundations for model design in artificial intelligence. In the human emotional generation process, priming effects and knowledge integration play pivotal roles. In cognitive science, priming effects refer to how external semantic or contextual cues alter the brain's processing of sensory stimuli. For instance, linguistic cues accelerate emotional categorization of ambiguous facial expressions. This process relies on the prefrontal cortex and hippocampus regulating visual pathways, enabling cross-modal semantic-sensory interaction \cite{bargh1999unbearable}. Simultaneously, the brain does not process emotions in isolation but relies on long-term semantic memory and contextual knowledge to interpret sensory inputs. The Conceptual Act Theory \cite{barrett2017theory} proposes that emotions are constructed through the integration of bodily signals with semantic knowledge via the hippocampus, default mode network, and prefrontal cortex, rather than being directly read. The hierarchical temporal processing mechanisms of the brain \cite{hasson2015hierarchical} and the information integration mechanisms between semantic memory and the prefrontal cortex \cite{binder2011neurobiology} provide interpretability for the bionic design of neurotransmitters. Inspired by this, we designed an algorithmic implementation of a complementary “dual-mechanism” cognitive framework: achieving embodied cross-modal emotion perception through cue-based expectation setting and knowledge integration.

\subsection{Dynamic Facial Expression Recognition}
Early studies on facial expression recognition relied on handcrafted features in controlled environments \cite{livingstone2018ryerson}. With the advent of deep learning and large-scale DFER datasets, data-driven approaches have become mainstream. Unlike static FER, DFER requires modeling spatiotemporal dynamics to capture expressive variations over time. The growing availability of in-the-wild datasets \cite{jiang2020dfew, wang2022ferv39k} has established DFER as a distinct research task, prompting the development of specialized methods to address its unique challenges. Recent advances in deep learning have shifted FER from static image analysis to motion-aware frameworks. Models like C3D \cite{tran2015learning} effectively capture spatiotemporal features and long-range dependencies, essential for modeling dynamic expressions. Meanwhile, vision-language models such as CLIP \cite{radford2021learning} and CLIPER \cite{li2024cliper} enable robust semantic alignment across modalities. Unlike their approach, ours focuses more on mining pre-trained textual prior knowledge to augment dynamic sentiment modeling from a semantically guided perspective.

\subsection{Dynamic Prompting with Cross-Domain Transfer}
Prompt learning \cite{lester2021power} has been extended from NLP to vision-language and vision-only models, enabling efficient adaptation to downstream tasks by learning task-specific prompts with minimal parameter updates. Recent text-based sentiment analysis leverages emotion-related prompts to reformulate classification as masked language modeling for data-efficient learning \cite{jim2024recent}, while visual prompt tuning methods such as CoOp \cite{zhou2022conditional}, VPT \cite{jia2022visual}, and MaPLe \cite{khattak2023maple} optimize learnable prompts to enhance cross-modal generalization in vision-language models like CLIP. However, these approaches mainly address coarse-grained object recognition and struggle with the fine-grained, temporally-evolving nature of DFER. To overcome this, we incorporate rich emotional textual prompts into facial imagery and introduce dynamic prompt tuning to capture temporal variations. Moreover, we enhance generalization via cross-domain knowledge transfer \cite{serrano2024knowledge}, which mitigates data scarcity and domain shifts (e.g., from controlled to in-the-wild settings through adversarial learning \cite{han2019adversarial} and feature disentanglement \cite{yang2022disentangled}. While prior methods rely on static FER as intermediates or knowledge distillation \cite{zhou2024ceprompt}, they offer limited gains and often suffer modal misalignment. In contrast, we propose a dynamic knowledge migration strategy that embeds emotional concepts into facial feature dynamics, aligning cross-modal semantics to improve real-world DFER performance on subtle and context-sensitive expressions.

\section{METHOD}

In DuSE design, we fully incorporate the complementary relationship between prompts and knowledge within the brain. The Hierarchical Temporal Prompt Cluster (HTPC) provides context-driven anticipatory expectations, simulating the process of modulating sensory processing sensitivity and the brain's hierarchical structure. The Latent Semantic Emotion Aggregator (LSEA) analogizes knowledge aggregation and emotional semantic processing, performing a posteriori construction to generate complete emotional concepts. DuSE's algorithmic implementation of a “dual-mechanism” neuroscience framework achieves embodied cross-modal emotion perception through pre-set expectations via prompts and knowledge-integrated construction.

\subsection{Preliminaries}
The overall architecture of DuSE is depicted in Fig. \ref{pip}. Specifically, the overall framework requires input downsampled video frame sequences $\mathcal{V}$ and enriched and expanded multi-class text descriptions $\mathcal{T}$. For the text part we have adopted the tagging combined with salient sentiment category features natural language description composition. They will be integrated as $\mathcal{V}_{in} \in \mathbb{R}^{t \times \mathcal{C} \times \mathcal{H} \times \mathcal{W}}$ and $\mathcal{T}_{in} \in \mathbb{R}^{c}$ as inputs to CLIP image encoder $\mathcal{F}(\cdot)$ and text encoder $\mathcal{G}(\cdot)$ under the shallow and deep prompting action of the HTPC. Where $t$ is the number of downsampled frames, $\mathcal{H}$, $\mathcal{W}$ and $\mathcal{C}$ are the information on pixel points of the image and $c$ is the number of categories to be categorized. The subsequently obtained video features $\mathcal{F}_{\mathcal{V}} \in \mathbb{R}^{t \times d}$ and text features $\mathcal{F}_{\mathcal{T}} \in \mathbb{R}^{c \times d}$, where d is the encoder output dimension of CLIP. Video features and text features after passing through the LSEA module will output the temporally modeled and semantically guided video fusion feature $\mathcal{V}_{g} \in \mathbb{R}^{d}$. Subsequently $\mathcal{V}_{g}$ will be aligned with $\mathcal{F}_{\mathcal{T}}$ and the contrast learning loss will be computed. Where $cls$ is the category corresponding to the correct label and $i$ traverses all categories.

\begin{equation}
\mathcal{L}=-\log \frac{\exp \left(\mathcal{V}_{g} \cdot \mathcal{F}_{\mathcal{T}}\left(cls\right)\right)}{\sum_i \exp \left(\mathcal{V}_{g}\cdot F_{\mathcal{T}}\left(i\right)\right)}
\end{equation}

\subsection{Hierarchical Temporal Prompt Cluster}
The prompt stream is not a single unit but a cluster of both shallow and deep prompts. The shallow prompts are designed for cross-modal interaction at the input stage, before the encoder, while the deep prompts are incorporated between layers of the encoder.

If we design $n$ learnable tokens and $\mathcal{M}$ prompt streams, where $\mathcal{M}$ cannot exceed the intrinsic number of layers $\mathcal{K}$ of the encoder, then the shallow prompting corresponds to when $\mathcal{M} = 1$, and the rest of the cases can be classified as deep prompting. The following two formulas can briefly summarize the process of prompting on the text side. Where $i = 1,2,...,\mathcal{M}$ represents the serial number of the layer affected by the prompt flow and $j = \mathcal{M}+1,\mathcal{M}+2,...,\mathcal{K}$ represents unaffected, both $\mathcal{P}_i^\mathcal{T}\in \mathbb{R}^{n \times d_\mathcal{T}}$ and $\mathcal{P}_j^\mathcal{T}\in \mathbb{R}^{n \times d_\mathcal{T}}$ are learnable tokens, while $\tau_i$ and $\tau_j$ are corresponding text encoder layers. $d_\mathcal{T}$ is the text encoder hidden layer feature dimension. The underscore “\_” in the following formulas denotes the output of the corresponding dimension, which our algorithm does not consider.

\begin{equation}
\left [ \mathcal{E}_i^\mathcal{T}, \_ \right ]=\tau_i\left ([ \mathcal{E}_{i-1}^\mathcal{T}, \mathcal{P}_{i-1}^\mathcal{T}\right ])
\end{equation}

\begin{equation}
\left [ \mathcal{E}_j^\mathcal{T}\right ]=\tau_j\left ([ \mathcal{E}_{j-1}^\mathcal{T}\right ])
\end{equation}

Correspondingly, the following two formulas can briefly summarize the process of prompting on the video side. Where both $P_i^V\in \mathbb{R}^{n \times d_V}$ and $P_j^V\in \mathbb{R}^{n \times d_V}$ are learnable tokens, while $\gamma_i$ and $\gamma_j$ are corresponding image encoder layers. $d_V$ is the image encoder hidden layer feature dimension.

\begin{equation}
\left[\mathcal{E}_i^\mathcal{V}, \_ \right]=\gamma_i\left([\mathcal{E}_{i-1}^\mathcal{V}, \mathcal{P}_{i-1}^\mathcal{V}\right])
\end{equation}

\begin{equation}
\left [\mathcal{E}_j^\mathcal{V}\right ]=\gamma_j\left ([ \mathcal{E}_{j-1}^\mathcal{V}\right ])
\end{equation}

In order to reflect the guiding role of textual semantics, the video-side learnable visual tokens are generated from their textual counterparts by the parameter-shared multi-layer perceptron $\mathcal{MLP}$. Where $\mathcal{W}$ is the parameter matrix, $b$ is bias parameters, and since it is a generalized regression task, no activation function is used before the output layer.

\begin{equation}
\mathcal{P}_i^\mathcal{V} = \mathcal{MLP}(\mathcal{P}_i^\mathcal{T})=\operatorname{ReLU}(\mathcal{W}\cdot \mathcal{P}_i^\mathcal{T}+b)
\end{equation}

In particular, due to the nature of the CLIP architecture as it applies to images, we would like to add dynamic prompts between frames. Therefore, for the first layer of prompts, after mapping the multilayer perceptron, a sinusoidal position encoding is performed in the frame-level dimension $t$, and then the position encoding vectors are summed up in the dimension of the number of learnable tokens $n$ for the broadcast mechanism.

With the HTPC, it is also possible to obtain $\mathcal{F}_{\mathcal{V}}$ and $\mathcal{F}_{\mathcal{T}}$ as described before. Shallow and deep prompts will be injected in a layered manner according to the hierarchy. For prompt depth, we define three strategies: $Shallow$ prompting strategy means affecting only the input layer, $Normal$ prompting strategy means affecting one-third of the encoder layers and $Deep$ prompting strategy means affecting two-thirds of the encoder layers. Since the performance met expectations and the large size of the CLIP ViT-L/14 model, $Deep$ prompting strategy was not applied to it. This is consistent with the subsequent ablation experiments.

\begin{equation}
\mathcal{F}_\mathcal{V} = \mathcal{F}\big(\mathcal{E}^\mathcal{V};\; \mathcal{P}_s^\mathcal{V}\!@1,\; P_d^V\!@{2..\mathcal{M}}\big)
\end{equation}

\begin{equation}
\mathcal{F}_\mathcal{T} = \mathcal{G}\big(\mathcal{E}^\mathcal{V};\; \mathcal{P}_s^\mathcal{T}\!@1,\; \mathcal{P}_d^\mathcal{T}\!@{2..\mathcal{M}}\big)
\end{equation}

\subsection{Latent Semantic Emotion Aggregator}
To effectively transfer the knowledge learned by CLIP to the expression recognition domain, we propose a text-guided knowledge transfer module to reduce the domain gap. This module leverages textual knowledge to guide visual feature learning, acting as a bridge that connects facial expression images with knowledge from the natural domain.

Since $\mathcal{F}_{\mathcal{V}} \in \mathbb{R}^{t \times d}$ is a multi-frame feature, we model its multi-frame fusion using a spatio-temporal split-attention mechanism. Since spatial information has already been taken into account in the CLIP image encoder, only time series modeling is performed here. This step is mainly realized based on the self-attention mechanism, where $\mathcal{Q}$, $\mathcal{K}$, and $\mathcal{V}$ represent the Query, Key, and Value matrices, respectively, and $d_k$ is the dimensionality factor for appropriate scaling. Subsequently passing it through a linear layer and applying temporal attention pooling to aggregate frame-level representations into a single fused feature $\mathcal{V}_o\in \mathbb{R}^{d}$. Where the weights $w_i$ are learned via via a learned scoring function followed by softmax over time. $w_i$ denotes the self-attention weight of the i‑th frame.

\begin{equation}
\text { Attention }(\mathcal{Q}, \mathcal{K}, \mathcal{V})=\operatorname{softmax}\left(\frac{\mathcal{Q} \mathcal{K}^T}{\sqrt{d_k}}\right) \mathcal{V}
\end{equation}

\begin{equation}
\mathcal{V}_m = \text{Linear}(\text{Attention}(\mathcal{F}_\mathcal{V},\mathcal{F}_\mathcal{V},\mathcal{F}_\mathcal{V})) \in \mathbb{R}^{t \times d}
\end{equation}

\begin{equation}
\mathcal{V}_o = \sum_{i=1}^{t} w_i \cdot \mathcal{V}_{m}^{(i)}, \quad \text{with} \sum w_i = 1
\end{equation}

Text features and video features are taken as inputs and measure their similarity by calculating the cosine of the vectors. Then, Softmax function is applied to normalize the similarity and construct the semantic vector $\mathcal{T}_o$ that is most similar to the image side, which is weighted and summed with the original features to achieve the latent feature embedding and obtain the text-guided image feature $\mathcal{V}_g$. In the specific implementation process, we introduce hyperparameter $\mathcal{N}$ to the generation process of $\mathcal{T}_o$, using the design of multiple heads. Specifically, we utilize a linear layer to map $\mathcal{V}_o$ and $\mathcal{F}_\mathcal{T}$ to $\mathcal{N}$ heads of the same dimension to obtain $\tilde{\mathcal{V}}_o^{(i)}$ and $\tilde{\mathcal{F}}_{{\mathcal{T}}}^{(i)}$, and perform the operations described earlier on these $\mathcal{N}$ pairs of features and finally average them to obtain $\mathcal{V}_g$.

\begin{equation}
\tilde{\mathcal{T}}_o^{(i)}=\operatorname{softmax} \left(\tilde{\mathcal{V}}_o^{(i)} \cdot \tilde{\mathcal{F}}_{{\mathcal{T}}}^{(i)}\right) \cdot \tilde{\mathcal{F}}_{{\mathcal{T}}}^{(i)}
\end{equation}

\begin{equation}
{\mathcal{V}_g}=\frac{1}{\mathcal{N}}\sum_i^\mathcal{N}\left({\beta \tilde{\mathcal{V}}_o^{(i)}+\left(1-\beta\right) \tilde{\mathcal{T}}_o^{(i)}}\right)
\end{equation}

The visual features provide low-level structural information for dynamic representation, while the semantic vectors contain high-level affective semantics from category-wide textual guidance. The weighting factor $\beta$ can balance the weights of the original visual information and the semantically guided information, and introduce enhancement through the semantic attention mechanism while preserving the visual details. The semantic bootstrapping mechanism in LSEA performs all-category soft bootstrapping via a multi-head semantic attention mechanism, which adaptively extracts and fuses the most relevant semantic vectors from all category texts for each visual feature. $\beta$ is used to control the influence of semantic guidance while attenuating potential noise from inter-class similarity.

\section{EXPERIMENTS}

\subsection{Experimental Setup}
\subsubsection{Datasets} Our study evaluates the effectiveness of DuSE on two in-the-wild, video-based DFER datasets: DFEW \cite{jiang2020dfew} and FERV39k \cite{wang2022ferv39k}. We use 5-fold cross-validation on DFEW to ensure thorough performance evaluation, and follow the predefined splits of FERV39k to remain consistent with its official protocol. This approach ensures a fair and rigorous comparison across different datasets and experimental settings. The results demonstrate DuSE’s robustness and accuracy, particularly in real-world conditions. 

\begin{table}[t]
\centering
\caption{Comparative results (\%). Our proposed DuSE performs well on both datasets for 7-class classification. Best results are in bold and second-best results are underlined.}
\scalebox{0.95}{
\begin{tabular}{ccccccc}
\toprule
\multirow{2}{*}{Method} & \multirow{2}{*}{Publication} & \multicolumn{2}{c}{DFEW} & \multicolumn{2}{c}{FERV39k} \\
\cmidrule(lr){3-4} \cmidrule(lr){5-6}
 & & UAR & WAR & UAR & WAR \\
\midrule
C3D \cite{tran2015learning} & CVPR'15 & 42.74 & 53.54 & 22.68 & 31.69 \\
P3D \cite{qiu2017learning} & ICCV'17 & 43.97 & 54.47 & 23.20 & 33.39 \\
I3D-RGB \cite{carreira2017quo} & ICCV'17 & 43.40 & 54.27 & 30.17 & 38.78 \\
3D ResNet18 \cite{tran2018closer} & CVPR'18 & 46.52 & 58.27 & 26.67 & 37.57 \\
EC-STFL \cite{jiang2020dfew} & MM'20 & 45.35 & 56.51 & - & - \\
Former-DFER \cite{zhao2021former} & MM'21 & 53.69 & 65.70 & 37.20 & 46.85 \\
NR-DFERNet \cite{li2022nr} & arXiv'22 & 54.21 & 68.19 & 33.99 & 45.97 \\
DPCNet \cite{wang2022dpcnet} & MM'22 & 57.11 & 66.32 & - & - \\
EST \cite{liu2023expression} & PR'23 & 53.94 & 65.85 & - & - \\
LOGO-Former \cite{ma2023logo} & ICASSP'23 & 54.21 & 66.98 & 38.22 & 48.13 \\
GCA-IAL \cite{li2023intensity} & AAAI'23 & 55.71 & 69.24 & 35.82 & 48.54 \\
MSCM \cite{li2023multi} & PR'23 & 58.49 & 70.16 & - & - \\
M3DFEL \cite{wang2023rethinking} & CVPR'23 & 56.10 & 69.25 & 35.94 & 47.67 \\
AEN \cite{lee2023frame} & CVPRW'23 & 56.66 & 69.37 & 38.18 & 47.88 \\
DFER-CLIP \cite{zhao2023prompting} & BMVC'23 & 59.61 & 71.25 & 41.27 & 51.65 \\
MAE-DFER \cite{sun2023mae} & MM'23 & 63.41 & 74.43 & 43.12 & 52.07 \\
EmoCLIP \cite{foteinopoulou2024emoclip} & FG'24 & 58.04 & 62.12 & 31.41 & 36.18 \\
SW-FSCL \cite{yan2024empower} & C\&C'24 & 57.25 & 70.81 & 36.83 & 49.87 \\ 
CLIPER \cite{li2024cliper} & ICME'24 & 57.56 & 70.84 & 41.23 & 51.34 \\
CDGT \cite{chen2024cdgt} & NN'24 & 59.16 & 70.07 & 41.34 & 50.80 \\
LSGTnet \cite{wang2024joint} & ASC’24 & 61.33 & 72.34 & 41.30 & 51.31 & \\
UMBEnet \cite{mai2024all} & MM'24 & \underline{64.55} & 73.93 & \textbf{44.01} & \underline{52.10} \\
\midrule
\textbf{DuSE(ours)} & - & \textbf{64.88} & \textbf{75.36} & \underline{43.39} & \textbf{53.05} \\
\bottomrule
\end{tabular}}
\label{table1}
\end{table}

\begin{table*}
\centering
\caption{Comparative results (\%) across different methods on various emotion categories in DFEW. Best results are in bold and second-best results are underlined.}
\scalebox{0.95}{
\begin{tabular}{ccccccccccccc}
\toprule
Method & Publication & Happy & Sad & Neutral & Angry & Surprise & Disgust & Fear & UAR & WAR \\
\midrule
C3D \cite{tran2015learning} & CVPR'15 & 75.17 & 39.49 & 55.11 & 62.49 & 45.00 & 1.38 & 20.51 & 42.74 & 53.54 \\
I3D-RGB \cite{carreira2017quo} & CVPR'17 & 78.61 & 44.19 & 56.69 & 55.87 & 45.88 & 2.07 & 20.51 & 43.40 & 54.27 \\
P3D \cite{qiu2017learning} & ICCV'17 & 74.85 & 43.40 & 54.18 & 60.42 & 50.99 & 0.69 & 23.28 & 43.97 & 54.47 \\
3D ResNet18 \cite{tran2018closer} & CVPR'18 & 76.32 & 50.21 & 64.18 & 62.85 & 47.52 & 0.00 & 24.56 & 46.52 & 58.27 \\
EC-STFL \cite{jiang2020dfew} & MM'20 & 79.18 & 49.05 & 57.85 & 60.98 & 46.15 & 2.76 & 21.51 & 45.35 & 56.51 \\
Former-DFER \cite{zhao2021former} & MM'21 & 84.05 & 62.57 & 67.52 & 70.03 & 56.43 & 3.45 & 31.78 & 53.69 & 65.70 \\
GCA-IAL \cite{li2023intensity} & AAAI'23 & 87.95 & 67.21 & 70.10 & 76.06 & 62.22 & 0.00 & 26.44 & 55.71 & 69.24 \\
SW-FSCL \cite{yan2024empower} & C\&C'24 & 88.35 & 68.52 & \underline{70.98} & \underline{78.17} & \underline{64.25} & 1.42 & 28.66 & 57.25 & 70.81 \\
LSGTnet \cite{wang2024joint} & ASC’24 & \underline{90.67} & \underline{71.70} & 70.48 & 76.71 & \textbf{65.01} & \underline{14.48} & \underline{40.24} & \underline{61.33} & 72.34 \\
\midrule
\textbf{DuSE (Ours)} & - & \textbf{92.89} & \textbf{81.05} & \textbf{72.76} & \textbf{78.51} & 62.69 & \textbf{20.69} & \textbf{45.57} & \textbf{64.88} & \textbf{75.36} \\
\bottomrule
\end{tabular}}
\label{table4}
\end{table*}

\begin{figure}[t]
  \centering
  \includegraphics[width=0.9\linewidth]{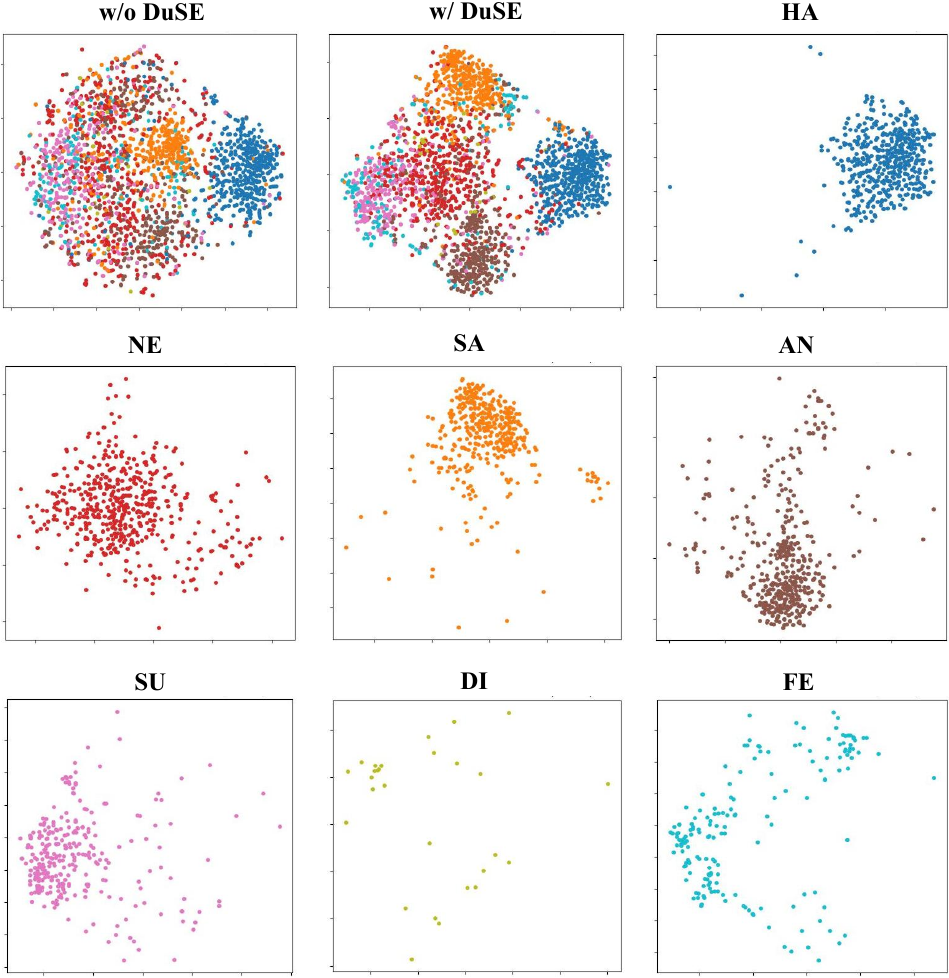}
  \caption{Global and category-specific t-SNE visualization of DuSE on DFEW-fold5. The clustering results show that DuSE has a significant enhancement effect.}
  \label{tsne}
\end{figure}

\begin{figure}[t]
  \centering
  \includegraphics[width=0.9\linewidth]{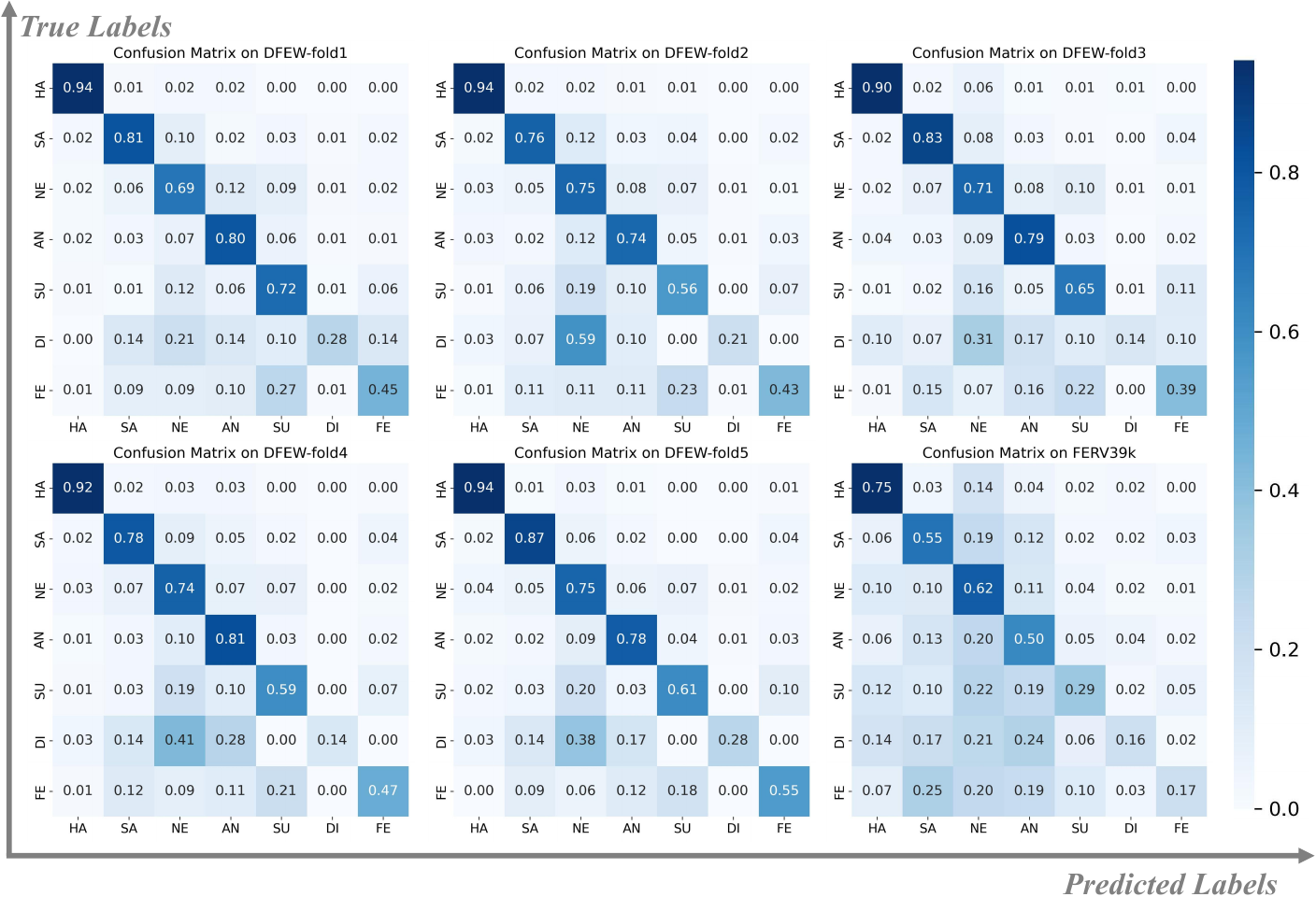}
  \caption{Confusion matrices of the DuSE on DFEW and FERV39k.}
  \label{mat}
\end{figure}

\subsubsection{Implementation details} For the visual input, all fixed 16-frame sequences in our DuSE experiments followed the sampling strategy described in related works and were resized to 224×224 pixels. To mitigate overfitting, we employed several data augmentation techniques, including random resized cropping, horizontal flipping, random rotation, and color jittering. The text section was designed with fixed prompt words that combine emotion categories with descriptions of subtle facial expression changes, paired with the learnable tokens mentioned earlier. These textual prompts provided semantic guidance for the model during training. 

All experiments were conducted in a high-performance computing environment equipped with 4 NVIDIA GeForce RTX 3090 GPUs. During training, we used the Adam optimizer with an initial learning rate of 0.001, and employed small-batch training with a batch size of 16. For the hyperparameters, we set the HTPC parameter $n$ to 4, the LSEA parameter $\mathcal{N}$ to 4 and $\beta$ to 0.7. To improve computational efficiency, we adopted automatic mixed-precision training, using half-precision floating-point numbers where applicable. This strategy accelerated the training process and reduced GPU memory usage, allowing for faster processing and more efficient scaling. Specifically, we conducted deployment and dynamic emotion recognition tests on an actual robotic head. DuSE is implemented as a pre-information sensing model for large multimodal large language models such as Qwen and LLaMA. The detected dynamic emotions are fed into the large model as part of the prompt. The entire framework can be deployed on the robot's head, utilizing the head camera to capture video data.

\subsection{Evaluation Metrics}
To evaluate model performance, we adopt two key metrics: Weighted Average Recall (WAR) and Unweighted Average Recall (UAR). WAR computes the average recall weighted by class sample size, making it suitable for imbalanced datasets where certain classes dominate. In contrast, UAR treats each class equally by averaging recall across all classes, regardless of their frequency, and is more appropriate for balanced datasets. Together, WAR and UAR provide a comprehensive assessment of model effectiveness and are widely used in the DFER field.

\subsection{Comparative Experiments}
In the context of our approach, experiments were conducted on two established DFER datasets, DFEW and FERV39k, with video data as the primary input. As shown in Table \ref{table1}, the DuSE method performs well on both datasets. Its consistently strong performance in terms of WAR and UAR metrics highlights the effectiveness of the method in achieving optimal results across multiple datasets. Compared to other CLIP-based methods, our approach demonstrates the advantage of cross-modal information interaction without the need for fine-tuning the encoder. Even when compared to methods such as MAE-DFER \cite{sun2023mae} and HiCMAE \cite{sun2024hicmae}, which are based on pre-training on a large amount of data, our method has some superiority. Table \ref{table4} demonstrates a comparison with previous work on the performance of DFEW in categorizing individual classes, where our method surpasses recent state-of-the-art methods and achieves the best results in essentially all classes.

Figures \ref{tsne} and \ref{mat} show the the t-SNE visualization during DFEW training and confusion matrix for the best DuSE results. It can be seen that the CLIP pre-trained model's own natural knowledge reserve allows for simple clustering of regular expression classes such as happy and sad, and the model's ability to perceive emotions such as anger and surprise is gradually enhanced as the training process proceeds and knowledge of the emotion domain is transferred.

\begin{table}[t]
\centering
\caption{Comparative results (\%) of the inter-module ablation experiments.}
\scalebox{0.9}{
\begin{tabular}{ccccccc}
\toprule
\multirow{2}{*}{\ \ HTPC\ \ } & \multirow{2}{*}{\ \ LSEA\ \ } & \multicolumn{2}{c}{DFEW} & \multicolumn{2}{c}{FERV39k} \\
\cmidrule(lr){3-4} \cmidrule(lr){5-6}
 & & UAR & WAR & UAR & WAR \\
\midrule
$\times$ & $\times$ & 55.64 & 66.80 & 33.74 & 46.26 \\
$\times$ & \checkmark & 58.86 & 70.28 & 35.77 & 47.85 \\
\checkmark & $\times$ & 60.27 & 71.83 & 36.53 & 48.80 \\
\midrule
\checkmark & \checkmark & \textbf{64.88} & \textbf{75.36} & \textbf{43.39} & \textbf{53.05} \\
\bottomrule
\end{tabular}}
\label{table3}
\end{table}

\subsection{Ablation Study}
In this work, we conduct ablation experiments on the DFEW and FERV39k datasets. This section focuses on intra-module and inter-module ablation experiments. The assessment metrics are consistent with the previous experiment.

Table \ref{table3} shows the results of the inter-module ablation experiments. Specifically, HTPC is replaced with the unimodal prompt tuning method CoOp, and LSEA is replaced with average pooling after ablation. The experimental results demonstrate that both modules are effective, with the introduction of each module individually significantly improving the model's performance. This highlights the importance of both the prompt and knowledge streams. Table \ref{table2} presents the intra-module ablation experiments in HTPC. We perform experiments on the number of learnable tokens and the prompt depth for each of the three CLIP pretrained models with varying specifications. The results demonstrate that our approach achieves improvements across baseline models of varying scales, with increasingly pronounced effects as the influence on the visual encoder intensifies. Table \ref{table5} shows the impact of hyperparameters on the experiment in LSEA. We mainly conducted ablation experiments on the number of heads $\mathcal{N}$ and fusion weight $\beta$. The results indicate that too few attention heads weaken the ability to capture semantic relationships across multiple categories, while too many may lead to overfitting or excessive learning of irrelevant features. The fusion parameter balances visual features with semantically enhanced features. An excessively high value weakens semantic guidance, causing the model to confuse fine-grained emotions, while an excessively low value results in overly strong semantic information and introduces noise from non-target categories.

\begin{table}[t]
\centering
\caption{Comparative results (\%) of the intra-module ablation experiments of HTPC (cross-modal prompt streaming).}
\scalebox{0.9}{
\begin{tabular}{ccccccc}
\toprule
\ \ Pre-trained model\ \ & \multirow{2}{*}{Strategy} & \multicolumn{2}{c}{DFEW} & \multicolumn{2}{c}{FERV39k} \\
\cmidrule(lr){3-4} \cmidrule(lr){5-6}
(parameters-frozen) & & UAR &  WAR & UAR & WAR \\
\midrule
\multirow{3}{*}{CLIP ViT-B/32} & $Shallow$ &  49.95 & 61.35 & 33.47 & 44.92 \\
 & $Normal$ & 53.22 & 64.70 & 35.02 & 46.87 \\
 & $Deep$ & 57.75 & 68.76 & 36.93 & 48.54 \\
\midrule
\multirow{3}{*}{CLIP ViT-B/16} & $Shallow$ & 55.27 & 65.97 & 35.14 & 46.13 \\
 & $Normal$ & 56.84 & 68.15 & 37.32 & 48.05 \\
 & $Deep$ & 59.86 & 71.94 & 38.95 & 50.03 \\
\midrule
\multirow{2}{*}{CLIP ViT-L/14} & $Shallow$ & 57.03 & 68.98 & 37.28 & 47.94 \\
 & $Normal$ & \textbf{64.88} & \textbf{75.36} & \textbf{43.39} & \textbf{53.05} \\
\bottomrule
\end{tabular}}
\label{table2}
\end{table}

\begin{table}[t]
\centering
\caption{Comparative results (\%) of the hyperparameter ablation experiments of LSEA (cross-domain knowledge streaming).}
\scalebox{0.9}{
\begin{tabular}{cccccc}
\toprule
\multirow{2}{*}{\ \ Hyperparameter\ \ } & \multirow{2}{*}{\ \ Value\ \ } & \multicolumn{2}{c}{DFEW} & \multicolumn{2}{c}{FERV39k} \\
\cmidrule(lr){3-4} \cmidrule(lr){5-6}
 & & UAR & WAR & UAR & WAR \\
\midrule
\multirow{3}{*}{$\mathcal{N}$} & 2 & 64.37 & 74.92 & 43.16 & 52.71 \\
 & 4 & \textbf{64.88} & \textbf{75.36} & \textbf{43.39} & \textbf{53.05} \\
 & 6 & 64.21 & 74.83 & 42.94 & 52.08 \\
\midrule
\multirow{4}{*}{$\beta$} & 0.3 & 61.26 & 71.82 & 38.55 & 48.73 \\
 & 0.5 & 62.01 & 73.56 & 40.98 & 50.09 \\
 & 0.7 & \textbf{64.88} & \textbf{75.36} & \textbf{43.39} & \textbf{53.05} \\
 & 0.9 & 63.38 & 74.33 & 42.47 & 51.74 \\
\bottomrule
\end{tabular}}
\label{table5}
\end{table}

\section{Conclusion}
This paper analyzes the gap between human dynamic emotion perception and existing DFER methods through cognitive affect theory. Inspired by the priming effect and knowledge integration in affect cognitive theory, we propose the Dual-Stream Semantic Enhancement (DuSE) framework. This framework integrates emotional concepts into facial appearance and leverages semantic information to transfer knowledge from general scenes to the data-scarce domain of facial expressions. This algorithmic implementation of a dual-mechanism cognitive science and neuroscience framework achieves embodied cross-modal emotion perception through prompting predefined expectations and knowledge integration. Extensive experiments on DFEW and FERV39k datasets validate our approach's effectiveness. The lightweight model framework serves as a deployment-friendly pre-processing emotion perception module for multimodal large language models. We will continue advancing research on agent emotion perception and expression, hoping this work provides valuable insights for other researchers.

\section*{Acknowledgments}

This work was supported by National Natural Science Foundation of China (No.62576109, 62072112, 62406075), National Key Research and Development Program of China (2023YFC3604802), Shanghai Key Technology R\&D Program (Grant No. 25511107200).

\bibliographystyle{IEEEtran}
\bibliography{aaai2026}

\clearpage

\onecolumn 

\begin{center}
    \vspace{2em}
    {\Large \textbf{Appendix}}
    \vspace{2em}
\end{center}


\section*{Additional Visualization}
\begin{figure}[ht]
    \centering
    \includegraphics[width=0.7\linewidth]{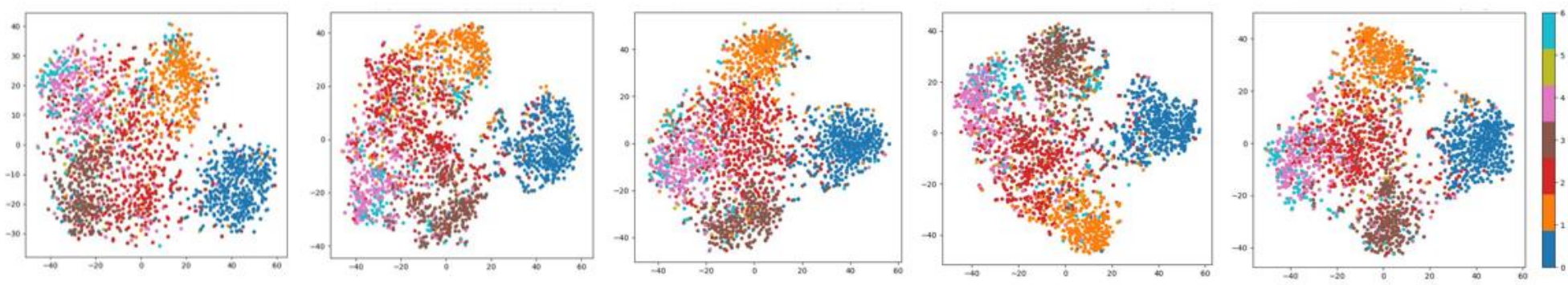} 
    \caption{t-SNE visualization on the DFEW 5-fold dataset.}
    \label{tsne1}
\end{figure}

We have supplemented DuSE’s t-SNE visualization results on DFEW in Figure \ref{tsne1} to demonstrate its overall performance on a real-world video dataset used for cross-validation.

\section*{Additional Deployment Details}
\begin{figure}[ht]
    \centering
    \includegraphics[width=0.4\linewidth]{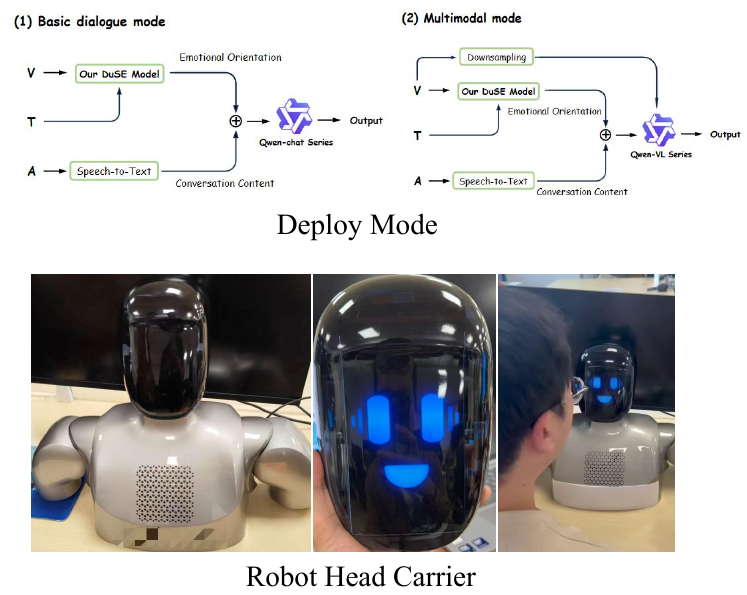}
    \caption{Deployment modes and implementation results, which are also mentioned in the demo video.}
    \label{deploy}
\end{figure}

Figure \ref{deploy} shows that as an easy-to-deploy emotion-sensing module, DuSE has been successfully integrated into a real robot head, using downsampled streaming video frames as dynamic visual input. Serving as a perceptual front-end for multimodal large language models such as the Qwen series, this framework seamlessly converts detected emotions into prompts that are fed into dialogue models, thereby enhancing the robot’s emotional perception capabilities during interactions. 

\end{document}